# ClipBot: an educational, physically impaired robot that learns to walk via genetic algorithm optimization

Diego Ulisse Pizzagalli[1,*], Ilaria Arini[2], Mauro Prevostini[2]

1. Faculty of Biomedical Sciences, USI – Switzerland
2. Faculty of Informatics, USI – Switzerland
* Correspondence: pizzad@usi.ch

## Abstract

*Educational robots allow experimenting with a variety of principles from mechanics, electronics, and informatics. Here we propose ClipBot, a low-cost, do-it-yourself, robot whose skeleton is made of two paper clips. An Arduino nano microcontroller actuates two servo motors that move the paper clips. However, such mechanical configuration confers physical impairments to movement.*

*This creates the need for and allows experimenting with artificial intelligence methods to overcome hardware limitations.*

*We report our experience in the usage of this robot during the study week 'fascinating informatics', organized by the Swiss Foundation Schweizer Jugend Forscht (www.sjf.ch). Students at the high school level were asked to implement a genetic algorithm to optimize the movements of the robot until it learned to walk. Such a methodology allowed the robot to learn the motor actuation scheme yielding straight movement in the forward direction using less than 20 iterations.*

**Keywords:** Educational robots, Genetic algorithms, Embedded artificial intelligence

# Introduction and results

Robots are by design interdisciplinary platforms that allow problem-based learning (PBL) of principles from mechanics, electronics, and informatics amongst others [1]. Moreover, the usage of robots in education is associated with increased motivation and satisfaction among students [2].

A variety of educational robot platforms are available, including simulators [3], open hardware robotic kits [4], and even multi-agent kits to experiment with swarm robotics [5].

During the study week 'Fascinating Informatics', *organized by the Swiss Foundation Schweizer Jugend Forscht (www.sjf.ch),* we proposed to three students the realization and the programming of ClipBot: a low-cost robot whose skeleton was made of two paperclips actuated by two servo motors powered by four AAA batteries (Fig. 1A).

An Arduino nano microcontroller with limited computing capabilities was used to control the movement of the servo motors by executing a specific control program (an ordered list of instructions) (Fig. 1B).

The main goal of the project assigned to the students was to make ClipBot capable to walk by optimizing the movements of the servo motors using artificial intelligence methods.

To this end, we employed a genetic algorithm [6] that iteratively generates control programs. These control programs are sequences of 15 instructions that the microcontroller executes (Fig.2, A), including movement of the front servo, movement of the back servo, and delays. Hence, it required only basic programming skills, allowed experimenting with array manipulation, and was compatible with the limited resources of the microcontroller.

The algorithm starts by generating 5 initial sequences with random commands. Then the robot executes the sequences and computes a fitness function, which is the displacement (in cm) towards the desired target. This was computed using an ultrasonic distance meter sensor on the head of the robot. Then, the algorithm selects the two sequences that obtained the maximum displacements and crosses them to generate a new sequence (child) (Fig. 2B).

To foster convergence of the algorithm, random mutations are included with a probability defined in the settings of the code. Child sequences that do not satisfy the existence conditions (i.e. delay < threshold, consecutive opposite commands) are discarded. Child solutions that respect the existence conditions are executed by the robot and their fitness function is evaluated. If a child solution has a fitness higher than the parents, then a parent gets removed. Sequences that made the robot fall were attributed with a fitness of -9.

To compensate for the limited amount of memory on the microcontroller, we introduced a catastrophe event [7], that randomly eliminates elements from the population regardless of their fitness function. The probability of this event and the number of elements to eliminate are customizable in the code settings.

Such a numerical scheme allowed ClipBot to learn to walk in less than 20 iterations, starting with a random population with an average traveled distance of -1.4 cm (going backward), and one individual in the population making the robot fall. After 20 iterations, the traveled distance increased up to 8cm in the forward direction (Fig. 3), with an absolute deviation angle of 10 degrees.
This demonstrates the applicability of genetic algorithms to optimize motor control in a way that makes ClipBot able to walk. The student mastered the genetic algorithm and proposed variations in line with the current literature (i.e. the carastrophe event) and successfully presented the project to the other participants of the study week.

## Discussion

Educational robotic platforms demonstrated suitability to use a PBL setting in high school education. Considering the pivotal role of machine learning in several fields such as medicine and engineering, it is important that students already at the high school level experiment with these technologies [8].

The proposed platform does not require advanced programming skills nor abilities with computation of derivatives which generally are acquired over multiple years [9]. Hence it provides a practical methodology to foster the education of students towards new technologies.

## Materials and Methods

ClipBot hardware was based on an Arduino Nano microcontroller v3. An Arduino nano expansion board was used to connect two SG90 servo motors.

Four AAA batteries connected in series were used to power the Arduino expansion board. A regulated 5V power supply was used for the microcontroller and the sensor, while a 6V unregulated supply from the batteries was used to directly power the servo motors.

An HC-SR04 ultrasonic distance meter was used to measure the traveled distance (fitness function).

A push button without latching was used to start a new iteration of the algorithm.

Code was written with Arduino IDE v 1.0.4 and is available under the GPL v3 Open Source license at
https://github.com/pizzagalli-du/clipbot/

## Author Contributions and Notes

D.U.P and I.A. designed the robot, and M.P. Organized the educational event and supervised. I.A. and D.U.P. wrote the software, D.U.P., I.A., and M.P. wrote the manuscript.

The authors declare no conflict of interest.

## Acknowledgments

We are thankful to the students of the study week Fascinating Informatics for their participation in this project. Benedikt Thelen for technical assistance.

## Fundings

This work was supported by the Swiss Foundation Schweizer Jugend Forscht **www.sjf.ch**

# Figures

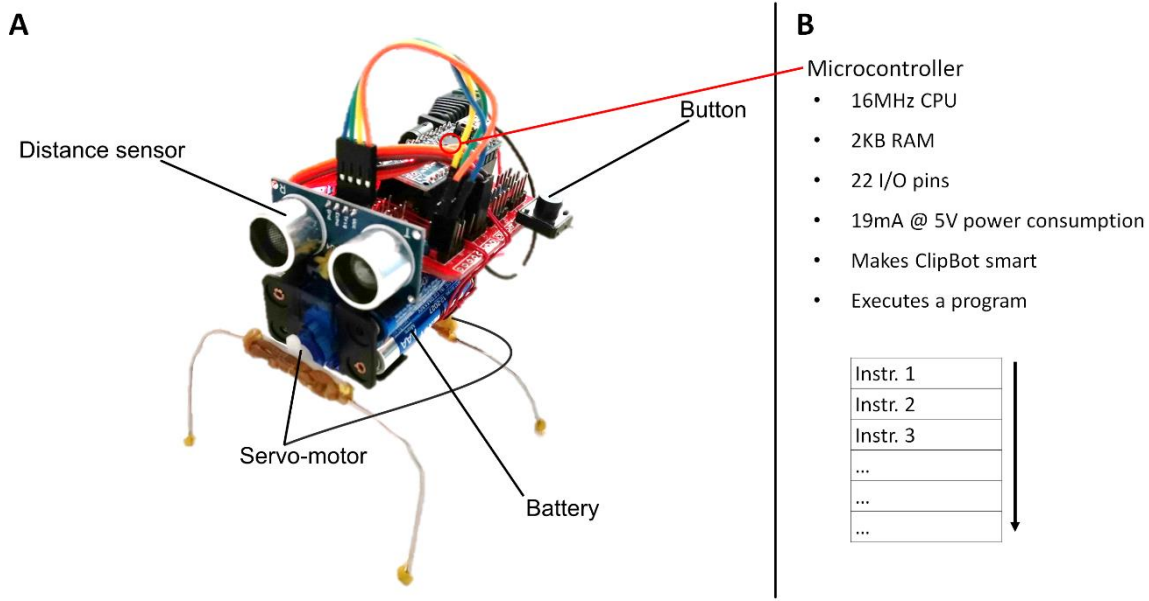

**Figure 1. Robot architecture.** A. ClipBot and its main components. B. Features of the microcontroller and control program

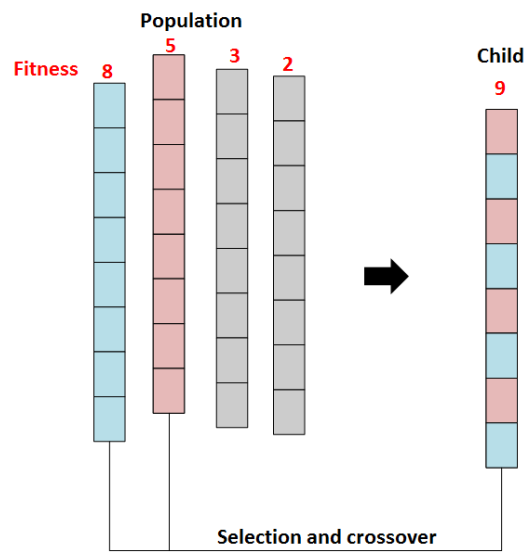

**Figure 2. Genetic optimization.** A. Structure of a single individual (control program) composed of 15 instructions encoded by a character and a number. B. Genetic algorithm that evaluates the fitness of each single individua, selects the best two and crosses them to generate a new individual.

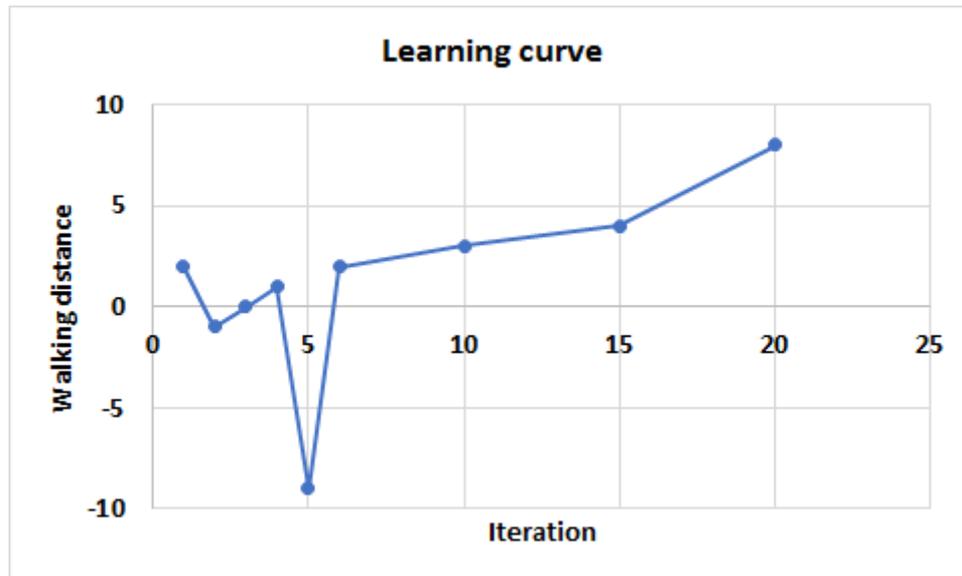

Figure 3. Comparison of the walking distance [cm] with respect to the iteration number